\documentclass[sigconf]{acmart}

\AtBeginDocument{%
  }

\setcopyright{acmlicensed}
\copyrightyear{2018}
\acmYear{2018}
\acmDOI{XXXXXXX.XXXXXXX}

\acmConference[Conference acronym 'XX]{Make sure to enter the correct
  conference title from your rights confirmation emai}{June 03--05,
  2018}{Woodstock, NY}
\acmISBN{978-1-4503-XXXX-X/18/06}

\usepackage{CJKutf8}

\usepackage[ruled,linesnumbered]{algorithm2e}
\usepackage{graphicx}
\usepackage{bbding}
\usepackage{float}
\usepackage{pifont}
\usepackage{subfig}
\usepackage{url}
\usepackage{booktabs}
\usepackage{amsmath}
\usepackage{bbding}
\usepackage{pifont}
\usepackage{wasysym}
\usepackage{utfsym}
\usepackage{fontawesome}
\usepackage{multirow}

\begin{document}

\title{APTNESS: Incorporating Appraisal Theory and Emotion Support Strategies for Empathetic Response Generation}

\author{Yuxuan Hu}
\authornote{Both authors contributed equally to this research.}
\email{huyx55@mail2.sysu.edu.cn}
\orcid{0009-0005-8571-118X}
\affiliation{%
  \institution{Sun Yat-Sen University}
  \city{Shenzhen}
  \state{Guangdong}
  \country{China}
}
\affiliation{%
  \institution{Shenzhen MSU-BIT University}
  \city{Shenzhen}
  \state{Guangdong}
  \country{China}
}
\author{Minghuan Tan}
\orcid{0000-0001-8287-0453}
\email{mh.tan@siat.ac.cn}
\authornotemark[1]
\affiliation{%
  \institution{Shenzhen Institute of Advanced Technology, Chinese Academy of Sciences}
  \city{Shenzhen}
  \state{Guangdong}
  \country{China}
}
\author{Chenwei Zhang}
\email{zhangchw7@mail2.sysu.edu.cn}
\orcid{0009-0000-6698-0652}
\affiliation{%
  \institution{Sun Yat-Sen University}
  \city{Shenzhen}
  \state{Guangdong}
  \country{China}
}
\affiliation{%
  \institution{Shenzhen MSU-BIT University}
  \city{Shenzhen}
  \state{Guangdong}
  \country{China}
}
\author{Zixuan Li}
\email{lizx76@mail2.sysu.edu.cn}
\orcid{0009-0002-5595-2343}
\affiliation{%
  \institution{Sun Yat-Sen University}
  \city{Shenzhen}
  \state{Guangdong}
  \country{China}
}
\affiliation{%
  \institution{Shenzhen MSU-BIT University}
  \city{Shenzhen}
  \state{Guangdong}
  \country{China}
}
\author{Xiaodan Liang}
\orcid{0000-0003-3213-3062}
\email{lizx76@mail2.sysu.edu.cn}
\affiliation{%
  \institution{Sun Yat-Sen University}
  \city{Shenzhen}
  \state{Guangdong}
  \country{China}
}
\author{Min Yang}
\orcid{0000-0001-7345-5071}
\email{min.yang@siat.ac.cn}
\authornote{Corresponding authors.}
\affiliation{%
  \institution{Shenzhen Institute of Advanced Technology, Chinese Academy of Sciences}
  \city{Shenzhen}
  \state{Guangdong}
  \country{China}
}

\author{Chengming Li}
\orcid{0000-0002-4592-3875}
\email{licm@smbu.edu.cn}
\authornotemark[2]
\affiliation{%
  \institution{Shenzhen MSU-BIT University}
  \city{Shenzhen}
  \state{Guangdong}
  \country{China}
}

\author{Xiping Hu}
\orcid{0000-0002-4952-699X}
\email{huxp@smbu.edu.cn}
\authornotemark[2]
\affiliation{%
  \institution{Shenzhen MSU-BIT University}
  \city{Shenzhen}
  \state{Guangdong}
  \country{China}
}

\renewcommand{\shortauthors}{Hu et al.}
\begin{CJK*}{UTF8}{gbsn}
\begin{abstract}
Empathetic response generation is designed to comprehend the emotions of others and select the most appropriate strategies to assist them in resolving emotional challenges. Empathy can be categorized into cognitive empathy and affective empathy. The former pertains to the ability to understand and discern the emotional issues and situations of others, while the latter involves the capacity to provide comfort. To enhance one's empathetic abilities, it is essential to develop both these aspects. Therefore, we develop an innovative framework that combines retrieval augmentation and emotional support strategy integration. Our framework starts with the introduction of a comprehensive emotional palette for empathy. We then apply appraisal theory to decompose this palette and create a database of empathetic responses. This database serves as an external resource and enhances the LLM's empathy by integrating semantic retrieval mechanisms. Moreover, our framework places a strong emphasis on the proper articulation of response strategies. By incorporating emotional support strategies, we aim to enrich the model's capabilities in both cognitive and affective empathy, leading to a more nuanced and comprehensive empathetic response. Finally, we extract datasets ED and ET from the empathetic dialogue dataset \textsc{EmpatheticDialogues} and ExTES based on dialogue length. Experiments demonstrate that our framework can enhance the empathy ability of LLMs from both cognitive and affective empathy perspectives. Our code is released at https://github.com/CAS-SIAT-XinHai/APTNESS.
 
\end{abstract}




\received{20 February 2007}
\received[revised]{12 March 2009}
\received[accepted]{5 June 2009}

\maketitle


\section{Introduction}
\begin{figure}[!htp]
\centering\includegraphics[width=\linewidth]{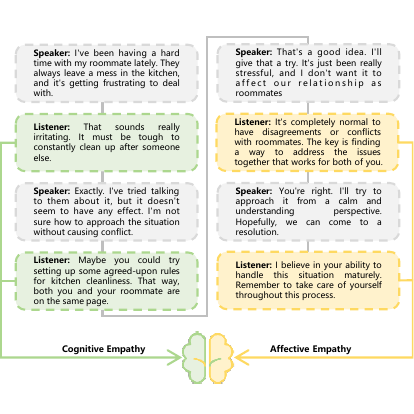}
\caption{ An example of empathetic response from ExTES dataset. Some empathetic responses are based on cognitive empathy, while others based on affective empathy.}
\label{motive}
\end{figure}

Empathy serves as a bedrock in human communication and emotional support, which is a critical component in empathetic response generation. In the field of sociology, empathy can be divided into two categories: cognitive empathy and affective empathy~\cite{reniers2011qcae}. 
Cognitive empathy refers to the ability to recognize and understand other people's emotions, without necessarily experiencing their feelings or circumstances firsthand~\cite{smith2006cognitive}. 
On the other hand, affective empathy involves forming a deep, heartfelt connection with another's emotions and situations, often through comforting, and responding to their emotions in a meaningful way~\cite{lawrence2004measuring}. 
The generation of empathetic responses necessitates the utilization of both cognitive and affective empathy skills, see Figure~\ref{motive}.

In contemporary society, the demand for emotional support and psychological solace is increasingly prevalent~\cite{keskin2014isn,rashkin-etal-2019-towards}. To meet this demand, artificial intelligence systems capable of engaging in empathetic dialogues have emerged as significant assets. These systems are expected to comprehend users' emotional situations and feelings, providing suitable responses to help users navigate their emotional difficulties and foster psychological comfort.

Several studies have begun to prioritize the distinction between cognitive empathy and affective empathy. For instance, semantic-based approaches, such as the one described in reference~\cite{sun2024rational}, strive to accurately understand others' emotions, thereby enhancing cognitive empathy. Methods based on COMET~\cite{hwang2021comet} aim to improve the emotional resonance of responses, thus enhancing affective empathy.
To further improve these two types of empathetic abilities in LLMs, we propose an empathetic response generation framework named APTNESS, which integrates appraisal theory and emotional support strategies. We believe that appraisal theory provides a cognitive framework for understanding how individuals interpret and respond to emotional stimuli. By incorporating this theory into LLMs, we can equip them with a wealth of external resources on empathetic expressive methods, thereby enhancing their ability to recognize and richly convey the empathy content of user inputs.
Furthermore, within the context of~\cite{liu-etal-2021-towards}, comforting an individual can employ different strategies, demonstrating a variety of conversational techniques. We believe that these diverse conversational techniques can assist the model in learning comprehensive empathetic capabilities, potentially leading to more effective and nuanced empathetic responses.

Specifically, to ensure the depth and breadth of the external resource we constructed, we introduce an empathetic emotional palette comprising 7 major categories and 23 subcategories of emotions. We then apply appraisal theory, as outlined in EmotionBench~\cite{huang2023emotionally}, to decompose these emotions. We follow the step-by-step evaluation process proposed in appraisal theory, breaking down emotions according to their constituent elements: the emotion itself, the factors influencing the emotion, and the situation. Leveraging ChatGPT's capabilities, we generate an external resource focused on empathetic responses.
To ensure that these responses effectively address both affective and cognitive empathy while providing richer language expressions, we use a two-stage empathetic response generation method. First, we facilitate dialogue to allow the LLM to generate preliminary responses. Next, we retrieve semantically similar empathetic responses from our external resource APT using semantic retrieval techniques. We then employ our fine-tuned strategy prediction module to gather strategic information. Finally, we combine all the information from these two stages to construct the final response.
To validate the effectiveness of our framework, we collect two types of empathetic dialogue data and conduct automated evaluations in a turn-based manner using GPT-4. Experiments with multiple LLMs show that our framework significantly enhances the model's empathetic abilities in both cognitive and affective empathy.

Our contributions are summarized as follows:
\begin{itemize}
\item We introduce an empathetic emotional palette to decompose emotions based on appraisal theory, creating a comprehensive empathetic response database.
\item We introduce APTNESS, a retrieval-augmented framework that utilizes appraisal theories and emotional support response strategies to enhance the empathetic capabilities of LLMs in generating empathetic responses.
\item We evaluate APTNESS using automated, turn-based methods with GPT-4 on both long and short empathetic dialogue datasets, ED and ET, demonstrating its effectiveness in affective and cognitive empathy.

\end{itemize}
\begin{figure*}[!htp]
  \centering\includegraphics[scale=1.2]{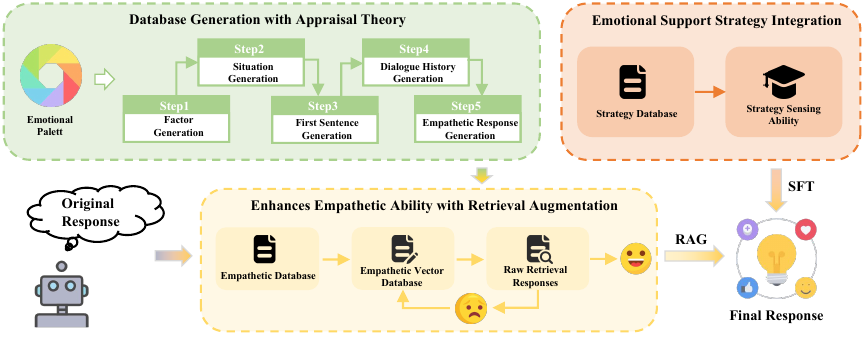}
\caption{The structure of the APTNESS framework is comprised of three core components: the generation of a empathetic response database with empathetic response appraisal theory, the retrieval augmentation module, and the integration of emotional support strategies module.}
\label{model}
\end{figure*}
\section{Related Work}
\subsection{Retrieval Augmented Generation}
With the advent of LLMS, generative language models have taken a prominent position in the AI landscape. Despite their advancements, these models often struggle with issues such as hallucinations, outdated information, and limited data, which can diminish their performance in specific scenarios. To mitigate these issues, retrieval augmented generation can be applied, providing external knowledge that helps these models correct errors and refine responses~\cite{gao2023retrieval}. The retrieval augment process primarily involves four key stages: query, encoding and retrieval, reranking, and generation. During the query stage, various methods are employed to enhance the richness of the query. Techniques like GRF and Query2doc~\cite{mackie2023generative,wang2023query2doc} expand queries by generating subtask documents or pseudo-documents using LLMs; Rewrite-Retrieve-Read~\cite{ma2023query}, on the other hand,  leverages transfer learning with smaller models to rewrite queries. The encoding and retrieval phase is the most crucial phase for retrieval augmented generation, directly influencing the quality of the retrieved text. Here, dense retrievers are commonly used. The dual-tower retriever training method introduced by DPR~\cite{karpukhin-etal-2020-dense} has gained significant traction, with extensions like Contriever~\cite{izacard2021unsupervised} applying dense retrievers to unsupervised data. The Tree of Clarifications~\cite{kim-etal-2023-tree} adapts retrieval processes using techniques such as breadth-first search tree traversal and pruning to refine document retrieval. In the reranking stage, methods such as GENREAD~\cite{yu2022generate} utilize clustering to select diverse prompts that guide the generation of varied external knowledge, while COMBO~\cite{zhang2023merging} employs a discriminator trained with silver labels to align generated documents with retrieved ones more effectively. Finally, in the generation phase, before the rise of LLMs, this process predominantly relied on a Reader specifically pretrained for retrieval enhancement, with models like Fid~\cite{izacard-grave-2021-leveraging} showing strong performance. However, with the emergence of powerful language generation models like ChatGPT and Llama~\cite{touvron2023llama}, the field has shifted towards using these models for generation through prompt-based methods, leveraging their superior language generation capabilities.

\subsection{Empathetic Response Generation}
The objective of the empathetic response generation task is to equip AI models with empathetic capabilities, enabling them to comprehend users' emotional needs and provide suitable emotional support. Numerous studies are being conducted on this task, with different research focusing on various aspects. Some research is centered on using scientific annotations and effective evaluations to aid in training empathetic response models. For instance, \textsc{EmpatheticDialogues}\cite{rashkin-etal-2019-towards} aims to enrich the corpus of empathetic responses by collecting short empathetic dialogues. ESConv\cite{liu-etal-2021-towards} is the first to introduce emotional support strategies into empathetic response tasks, enhancing conversational skills. Building on ESConv, ExTES~\cite{zheng2023building} redefines emotional support strategies, using ChatGPT to generate a more diverse emotional support strategy through self-chat. This results in longer and more diverse conversations in the corpus. Following the emergence of the COMET ~\cite{hwang2021comet} model pretrained on the commonsense corpus, some studies are beginning to focus on integrating external common knowledge into empathetic response tasks. For example, \cite{sabour2022cem} enriches the common sense information in historical dialogues using COMET, marking the first introduction of external common knowledge into the task. The Global-to-Local Hierarchical Graph Network \cite{peng2022control} utilizes graph network models, considering local common knowledge. CASE \cite{zhou-etal-2023-case} tries to build upon COMET and ConceptNet \cite{speer2017conceptnet} and align the user's cognition and affection at both coarse-grained and fine-grained levels with a language model. In the latter part of the ESConv research, some studies are beginning to investigate the integration of emotional support strategies. For example, MISC~\cite{tu-etal-2022-misc} attempts to focus on including fine-grained emotional understanding and mixed strategies to enrich semantic information in the text. SUPPORTER~\cite{zhou-etal-2023-facilitating} focuses on the strategy of multi-turn dialogues, using reinforcement learning and introducing coherence rewards and emotional support to guide strategy learning. However, with the rise of LLMs, there has been no research on integrating emotional support strategies into empathetic response dialogues, which is the primary focus of our research. Research on empathetic response has primarily focused on semantics with LLM. For example, Lamb~\cite{sun2024rational} aims to enhance the emotional cognition of models by studying the use of rational reasoning and emotional expression. The method ~\cite{qian-etal-2023-harnessing} explores the impact of external empathetic knowledge, semantically similar in-context learning, and multi-step thinking on LLMs. In contrast to their research, we propose a framework for empathetic response LLMs that integrates retrieval augmentation and emotional support strategy based on an empathetic appraisal theory. This approach aims to fill the gap in the current research landscape, offering a new perspective on empathetic response tasks in the era of LLMs.

\section{Methodology}
Inspired by appraisal theory, we first introduce a comprehensive empathetic emotional palette and decompose it into situations to create the empathetic response database, APT. We then enhance the empathetic response capabilities of LLMs using retrieval augmentation and strategy integration methods. The structure of the APTNESS framework is depicted in Figure~\ref{model}.

\subsection{Task Formulation}
Formally, given a dialogue history, which comprises alternating utterances between the Speaker and the Listener, defined as $C=\{ S_{1},  L_{1},  \allowbreak S_{2}, L_{2}, ..., S_{N_i-1}, L_{N_i-1},S_{N_i} \}$ of $2N_i-1$ utterances, where $S_{i}$ and $L_i$ denote the i-th utterance of the Speaker and the Listener, and $N_i$ represents the total number of utterances of one person in dialogue $i$. Our aim is to provide an empathetic response, denoted as $R$, corresponding to $L_{N_i}$.

\subsection{The APT Database}

This section describes how we created the Empathetic Response Database APT based on appraisal theory. The database generation procedure is illustrated in the green part of Figure~\ref{model}.
\subsubsection{Appraisal-Theory-based Emotion Decomposition}

Here, we provide a brief overview of the emotional support evaluation theory included in EmotionBench~\cite{huang2023emotionally}. EmotionBench proposes assessing the empathetic capabilities of LLMs by examining the fluctuation of emotions in specific situations. The creators assembled a dataset, EmotionBench, comprising over 400 different situations, each corresponding to one of eight negative emotions. Additionally, they identified 36 influencing factors across these situations, providing a comprehensive framework for evaluating the empathetic responses of LLMs

We believe that the process of decomposition mirrors how humans empathize in daily life: initially considering the emotions of the individual seeking assistance, then reflecting on the factors influencing these emotions, and taking into account the individual's circumstances before formulating a response. This process aligns with EmotionBench's approach to evaluating the emotional support capabilities of LLMs. Therefore, to enhance their capacity to emulate human understanding during empathetic response generation, LLMs must incorporate this process of emotional decomposition when responding to different situations.

\begin{figure}[!htp]
\centering\includegraphics[scale=0.15]{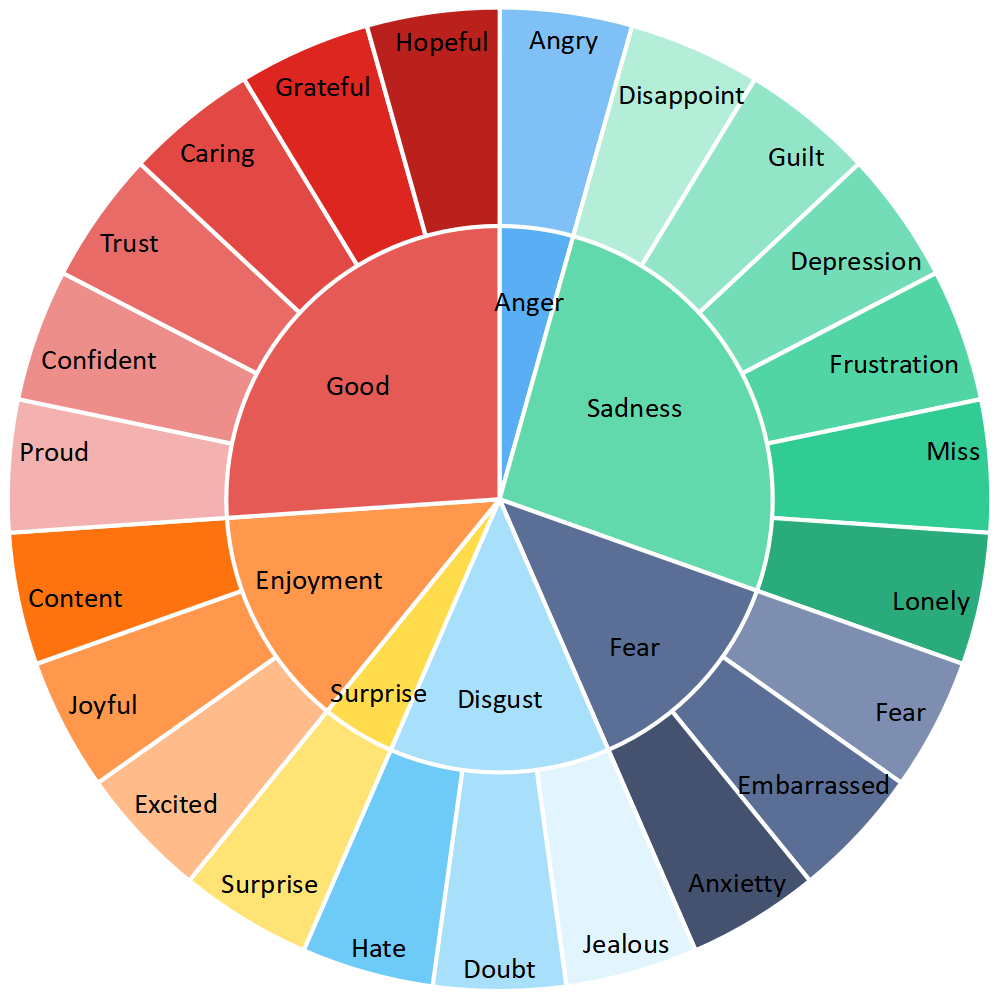}
\caption{Empathetic Emotional Palette}
\label{Empathetic Emotion Palette}
\end{figure}
\subsubsection{Emotional Palette}

To guarantee the diversity and comprehensive coverage of emotions in our empathetic response database, we assert that it should include a broad spectrum of emotions. Given the complexity and variability of human emotions, coupled with our incomplete and evolving understanding of them, there is currently no universally accepted standard for emotion categories. Existing emotion categories range from 4, 6, 8, to even over a dozen or more categories. Drawing on empathetic dialogue research \cite{rashkin-etal-2019-towards} and previous emotion classification literature \cite{ekman1993facial, Xu2007Constructing}, we develop a comprehensive empathetic emotional palette. This palette includes 7 main categories and 23 subcategories, as detailed in Figure~\ref{Empathetic Emotion Palette}. We adopt the seven main categories from \cite{Xu2007Constructing} and further refine them to cater to empathy-specific situations. Additionally, to enrich the spectrum of positive emotions and merge some emotions in empathetic contexts, we propose 23 subcategories.

\subsubsection{Construction}

Upon analyzing empathetic dialogues in the \textsc{EmpatheticDialogues}~\cite{rashkin-etal-2019-towards} dataset, we observe that in short conversations, such as those with only two turns, the Listener often provides empathetic responses in the final turn. Consequently, we decide to generate empathetic responses using short dialogues. We construct our database using ChatGPT\footnote{gpt-3.5-turbo} in two stages. All prompts we used for construction are included in our code.
\begin{enumerate}
    \item In the first stage, we sequentially generate emotional factors and situations for all emotions in our emotional palette using few-shot examples from EmotionBench~\cite{huang2023emotionally}.
    \item In the second stage, we use the emotional palette, along with the generated factors and situations, to prompt ChatGPT to generate short dialogues with empathetic responses. During the generation process, we encounter issues of repetitive generation when repeatedly prompting ChatGPT to generate empathetic dialogues. Inspired by the significant role of thought chains in LLM reasoning, we decide to have ChatGPT generate dialogues in a step-by-step manner: first generate the initial utterance of an empathetic dialogue; then continue the dialogue to generate the whole empathetic conversation; finally, rethink the emotion, factor, and situation of the dialogue, and regenerate the last turn of the listener with an empathetic response.
\end{enumerate}

\paragraph{Statistics} After the generation process, our final APT Database is composed of 7 major emotion categories, 23 emotion subcategories, 230 factors, 2,415 situations, and 9,663 dialogues. This comprehensive dataset ensures wide coverage of empathetic scenarios and emotional responses, providing a robust foundation for retrieval augmentation in empathetic response generation systems.
\subsection{Two-Stage Empathetic Response Generation}

Drawing inspiration from recent advancements in retrieval augmentation for LLMs, along with research on emotional support strategies for generating empathetic responses, we propose a two-stage empathetic response generation method. This method consists of two main components: empathetic response retrieval and emotional support strategy integration. The method is shown in the yellow and orange parts of Figure~\ref{model}.

\subsubsection{Empathetic Responses Retrieval}

Merely providing empathetic responses based solely on dialogue history is insufficient for LLMs. This is because the semantic information in the dialogue history is limited, and LLMs can only rely on their conversational abilities to provide responses. Empathetic response generation systems require more external resources to assist LLMs in generating empathetic responses. LLMs complete pre-training with a vast amount of data, embedding diverse forms of knowledge in their weights. Therefore, it is crucial to maximize the empathetic capabilities of these LLMs in the empathetic response generation task. Furthermore, our experiments show that although LLMs exhibit remarkable expressive capabilities, they tend to over-rely on offering suggestions during empathetic response generation, as shown in Table~\ref{tab
study}. This approach does not align with our expectations, as empathetic responses should display diversity.

Inspired by recent work on retrieval augmentation in LLMs, we aim to enhance the empathetic response capabilities of LLMs using our constructed empathetic response resource, APT. By leveraging this comprehensive external resource, we aspire to stimulate both the affective and cognitive empathetic capabilities and rich response expression of LLMs. We adopt a vectorized semantic retrieval approach, which can generate empathetic responses that resonate with similar emotions and situations. This approach aids LLMs in acquiring empathetic knowledge, thereby improving their ability to provide meaningful and empathetic responses.

Specifically, we utilize the dialogue response $G_1$ generated by LLMs for retrieval. The retrieval model we employ is Nomic Embed~\cite{nussbaum2024nomic}, which is trained on a vast amount of text data. This model uses contrastive learning objectives to map sentences into a vector space. The semantic similarity between two responses is then calculated by comparing their respective vector representations. This method allows us to accurately measure the semantic closeness of different responses, thereby facilitating more effective retrieval of empathetic responses：
\begin{align}\label{eqn1}
	G_1 &= \mathit{LLM}(C) \\
	E_{G_1} &= \mathit{Enc}(G_1) \\
	E_{\mathit{Database}} &= \mathit{Enc}(\mathit{Database}) \\
	R_1, R_2, R_3,...,R_{K} &= \mathit{TopK}(\frac{E_{G_1 } \cdot E_{\mathit{Database} }}{\left|E_{G_1 }\right|\left|E_{\mathit{Database} }\right|})
\end{align}
where $Enc$ represents the sentence encodings of the response, and $TopK$ selects the $K$ responses with the highest similarity between two sentence vectors.

\subsubsection{Emotional Support Strategy Integration}
Emotional support strategies represent distinct conversational skills used at each stage of a dialogue. Previous research has demonstrated that incorporating these strategies can effectively guide models in generating empathetic responses~\cite{zheng2023building,liu-etal-2021-towards}. Upon examining the definition of emotional support strategies, we find that the techniques they cover are broad. These strategies can primarily be divided into those aimed at understanding the emotions and situations of the help-seeker, and those focused on providing empathetic assistance. This division corresponds precisely to cognitive empathy and affective empathy. Therefore, we believe that incorporating emotional support strategies can help LLMs develop the skills relevant to both types of empathy, thereby enhancing their empathetic abilities.

\begin{figure}[!h]
\centering\includegraphics[width=\linewidth]{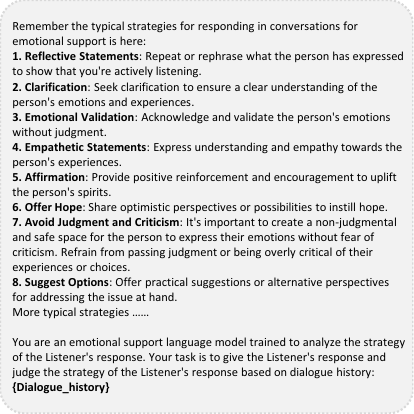}
\caption{Prompt for strategy integration LoRA SFT.}
\label{Prompt_sft}
\end{figure}

Therefore, our objective is to introduce emotional support strategies as external guidance for LLMs in generating empathetic responses. We believe that strategies should be incorporated only when needed; when not required, the LLM can generate responses according to its own process. To achieve this, we train a LoRA~\cite{hu2021lora} module for strategy prediction within LLMs using datasets labeled with empathetic strategies, such as ESConv~\cite{liu-etal-2021-towards} and ExTES~\cite{zheng2023building}. Our analysis reveals that most of the data in the 'Others' category of ESConv consist of greetings or closing phrases, prompting us to rename this category as 'Greetings.' Similarly, statements in ExTES without defined strategies also exhibit a consistent theme, leading us to label them as 'Greetings.' In comparison, ExTES contains 17 finely annotated strategies, whereas ESConv has only 8. Furthermore, while ExTES generally has one response corresponding to one strategy, some responses in ESConv correspond to multiple strategies. The prompt we defined for the strategy prediction LoRA SFT is shown in Figure~\ref{Prompt_sft} and includes three parts: strategy definition, task definition, and dialogue history.

\begin{algorithm}[t]  
\caption{Procedure of the Two-Stage Empathetic Response Generation} 
\label{alg1}
\KwIn{
 A dialogue history $C$; \\
\qquad\quad \ Empathetic response database APT Database $\mathit{APTD}$;\\
\qquad\quad \ Emotional support strategy prediction dataset $D_S$;\\
\qquad\quad \ Emotional support strategy definitions $f_{def}$\\
} 

\SetKwInOut{Output}{Initialize}
\Output{Semantic retriever $\mathit{Retriever}$}
\KwOut{LLM's response $R_{\mathit{final}}$}

\ForEach{iter}{
\textbf{Retrieve Responses:}\\
Generate original response: $G_1=LLM(C)$\\
Retrieve semantically similar response from APTD: $R_1,R_2, R_3,...,R_{K} = \mathit{Retriever}(G_1)$;\\
Response set: $R=[R_0(G_1),R_1,R_2, R_3,...,R_{K}]$;\\
Responses' dialogue history set from APTD: $H=[H_0(C),H_1,H_2, H_3,...,H_{K}]$;\\
\textbf{Strategy Integration:}\\
Supervised fine-tuning a strategy prediction LoRA module: $LLM_{LoRA}=SFT_{LoRA}(D_S)$;\\
\For{$H_i$ in $H$}{
    {Predict strategy: $S_i=\{s_{i1},s_{i2},...s_{il}\}=LLM_{LoRA}(H_i)$}
}
Strategy set: $S=[S_1,S_2, S_3,...,S_{K}]$;\\
Deduplication：$S=[S_1,S_2, S_3,...,S_{K'}]$;\\
Strategy definition set: $S_{def}=[f_{def}(S_1),f_{def}(S_2), f_{def}(S_3),...,f_{def}(S_{K'})]$;\\
\textbf{Generate Final Response:}\\
$R_{\mathit{final}}=\mathit{Prompt}(C,R,S,S_{def})$
}
\end{algorithm}
\subsubsection{The Augmentation Procedure}
Our framework combines retrieval augmentation with the integration of emotional support strategies: In the first stage, we let the LLM generate a response $G_1$ of the dialogue history $C$, and then retrieve semantically similar empathetic responses $R_1, R_2,...,R_{K}$ and their corresponding dialogue history $H_1, H_2,...,H_{K}$. In the second stage, we enhance the LLM's empathetic capabilities by using the emotional support strategy integration module to predict appropriate responses' strategies $S_1,S_2, ...,S_{K}$ for all dialogue histories $C,H_1,H_2, H_3,...,H_{K}$. And a single dialogue history's response may have multiple strategies. 
Next, we deduplicate it to obtain $S_1,S_2,...,S_{K'}$ and find the corresponding strategy explanations $f_{def}(S_1),f_{def}(S_2),...,f_{def}(S_{K'})$. The description of the whole procedure is as shown in Algorithm~\ref{alg1}. 

\begin{figure}[!htp]
\centering\includegraphics[width=\linewidth]{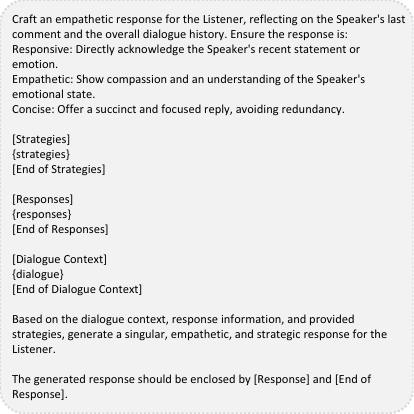}
\caption{Prompt for the two-stage empathetic response generation}
\label{Prompt}
\end{figure}

The prompt we design is as Figure~\ref{Prompt}. We place the dialogue history $C$ in the '{dialogue}' section of the prompt. The generated response $G_1$ and retrieved responses $R_1, R_2,...,R_{K}$ are sequentially placed in the '{responses}' section of the prompt in the format [Response K] $R_K$ [End of Response K]. The strategies $S_1,S_2, S_3,...,S_{K}$, generated by the strategy injection module, are deduplicated and ordered to become $S_1,S_2, S_3,...,S_{K'}$. These are then placed in the '{strategies}' section of the prompt in the format [Strategy $K'$] $S_{K'}$, which is defined as $f_{def}(S_{K'})$[End of Strategy $K'$].

\subsection{Automatic Evaluation of LLM-based Empathetic Response Generation}
\subsubsection{Evaluation Metrics}
In the domain of empathetic response generation, the quality assessment of multi-turn conversations still relies on traditional metrics such as BLEU-n and Dist-1/2~\cite{qian-etal-2023-harnessing,sun2024rational,tu-etal-2022-misc}, which benchmark against real labels. We argue that evaluating LLMs using these metrics is somewhat limited, as the complexity and comprehensiveness of the responses these models produce may significantly deviate from real labels. Consequently, evaluating multi-turn empathetic response generation without the involvement of human experts presents a challenge.

\begin{figure}[!htp]
\centering\includegraphics[width=\linewidth]{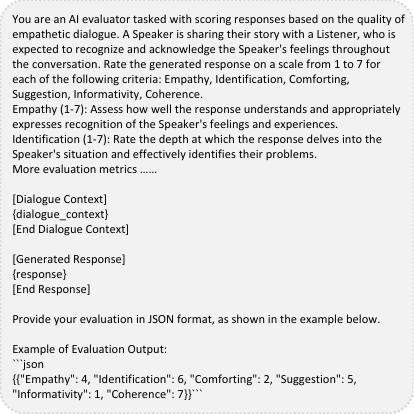}
\caption{Prompt for the automatic score.}
\label{evaluation prompt}
\end{figure}

To tackle this, we incorporate human evaluation metrics from previous research.
Human evaluation metrics from previous research divide into two categories: Empathy, Coherence, Informativity, and Fluency~\cite{qian-etal-2023-harnessing,sun2024rational,rashkin-etal-2019-towards}, as well as Fluency, Identification, Comforting, Suggestion, and Overall~\cite{liu-etal-2021-towards,zhou-etal-2023-facilitating}.
We select Empathy (Assess how well the response understands and appropriately expresses recognition of the Speaker's feelings and experiences.) as our primary evaluation metric, while Coherence (Evaluate the relevance and logical connection of the response to the dialogue context.), Informativity (Determine the richness and value of the information provided in the response.), Identification (Rate the depth at which the response delves into the Speaker's situation and effectively identifies their problems.), Comforting (Score the proficiency of the response in providing comfort and support.), and Suggestion (Rate the quality of the suggestions offered for addressing the Speaker's issues.) serve as our sub-metrics. Inspired by \cite{komorita1965number}, emotional evaluation metrics are graded on a 7-point scale. To obtain reliable results automatically, we use GPT-4 to assign scores to responses on this scale. The evaluation prompt is illustrated in Figure~\ref{evaluation prompt}."

\subsubsection{Turn-based Empathetic Response Evaluation}
In this work, we adopt turn-based empathetic response evaluation. Specifically, for each dialogue $C_i$ with $N_i$ turns, we use all conversation history till the $j$-th turn plus the query from the Speaker, denoted as $C_{ij}$, together with the generated the response $R_{ij}$ from LLMs. 
When scoring, GPT-4 assigns a score based on a dialogue history $C_{ij}$ and its response $R_{ij}$.
The scoring process is denoted as a function $\textit{score}$. 
Then we average all scores obtained for a complete dialogue as the score of an whole dialogue. 
Finally, we average all dialogues' scores as the final score $SC$, the equations are shown as follows:


\begin{align}\label{eqn2}
    SC=\frac{1}{N} \sum_{i=1}^{N}\left(\frac{1}{N_i} \sum_{j=1}^{N_i} \textit{score}(C_{ij}, R_{ij})\right)
\end{align}
 where $N_i$ is the number of turns for the $i$-th dialogue and $N$ is the number of dialogues.

\section{Experiments}

\subsection{Setup}

\paragraph{Datasets.}
We select the longest dialogues from the \textsc{EmpatheticDialogues} test set and the ExTES dataset to constitute our test sets, which we respectively refer to as ED and ET.
\begin{itemize}
    \item \textbf{ED}. From the \textsc{EmpatheticDialogues} dataset~\cite{rashkin-etal-2019-towards}, we extract 30 dialogues, each with 4 turns, forming a short dialogue test set of 120 turns in total.
    \item \textbf{ET}. We select 10 dialogues from the ExTES ~\cite{zheng2023building} dataset to form the long dialogue test set. Each dialogue in this set encompasses 12 turns, also totaling 120 turns. 
\end{itemize}

\paragraph{Models.} In our experiments, we select Llama2-7B\footnote{\url{https://huggingface.co/meta-llama/Llama-2-7b-chat-hf}}, Llama3-8B\footnote{\url{https://huggingface.co/meta-llama/Meta-Llama-3-8B-Instruct}}, and ChatGPT\footnote{gpt-3.5-turbo-16k-0613} as our baseline models, GPT4\footnote{gpt-4} as our automatic evaluation model. We set the temperature to 0.95 and $top-p$ to 0.7 to ensure the richness of the responses and to follow the specified format of our instruction output, which helps reduce errors in following instructions.

We use the Nomic Embed encoder~\cite{nussbaum2024nomic}, specifically version 1.5 from the Ollama library\footnote{\url{https://ollama.com/}}, and employ the LlamaIndex library\footnote{\url{https://www.llamaindex.ai/}} to construct our Information Retrieval (IR) system. The APT database, which we previously built, serves as the retrieval database. Given our primary focus on empathetic response generation, we extract all responses from the database, totaling 19,896 entries. To avoid exceeding the token limit for LLM input, we set the top-k value to 2.

For the experiments involving the fine-tuning of the strategy integration module, we select 10,000 data samples from the ExTES and ESConv datasets, while maintaining the proportion of data for each strategy. Due to the considerable variation in data volumes across different strategies in ExTES, we initially extract 100 data samples from each strategy with fewer data points to address the long tail problem. The models are fine-tuned for 5 epochs with a learning rate of 1e-5 and a batch size of 64. During fine-tuning, we combine the queries and dialogue histories with the prompt, as illustrated in Figure~\ref{Prompt_sft}.

\paragraph{Metrics.} In our experiments, the evaluation metrics used are Empathy (Emp.), Coherence (Coh.), Informativity (Inf.), Identification (Iden.), Comforting (Comf.), and Suggestion (Sug.). Among these, Empathy is the primary evaluation metric, while the others serve as sub-metrics. We employ a turn-based methodology for automatically evaluating empathetic response generation. The GPT-4 automatic evaluation score scale for all metrics ranges from 1 to 7. The evaluation prompt is shown in Figure~\ref{evaluation prompt}.
\begin{table*}[h] 
    \centering
    \caption{Comparison of different LLMs with APTNESS performance. The maximum value for each evaluation metric is bolded, and the second highest value is underlined.}
    \begin{tabular}{ccccccccccccccc}
    \toprule
     \multirow{3}{*}{Foundation Model}&\multirow{3}{*}{Method} & \multicolumn{6}{c}{ED}&&\multicolumn{6}{c}{ET}\\
     \cmidrule{3-8}\cmidrule{10-15}
     &&Emp.&	Coh.	&Inf.&	Iden.&	Comf.&	Sug.  && Emp.&	Coh. &Inf.&	Iden.&	Comf.&	Sug.        \\
     \midrule
     \multirow{3}{*}{Llama2-7B}&GEN&5.56&6.40&\textbf{4.76}&4.75&4.89&\textbf{4.53}& &6.06&6.62&\textbf{5.48}&5.16&5.98&\underline{5.82}\\
     &RAG&6.08&6.41&4.22&5.13&\underline{5.51}&4.10&	&\underline{6.45}&\underline{6.63}&\underline{5.15}&\underline{5.65}&\underline{6.40}&\underline{5.82}\\
     &APTNESS&\underline{6.22}&6.46&\underline{4.62}&\underline{5.20}&\textbf{5.63}&\underline{4.39}&	&\textbf{6.50}&6.51&\textbf{5.48}&\textbf{5.72}&\textbf{6.46}&\textbf{6.03}\\\midrule
    \multirow{3}{*}{Llama3-8B}&GEN&5.72&\underline{6.62}&4.17&4.69&5.02&3.87&	&5.99&\textbf{6.72}&4.44&5.10&5.73&4.97\\
     &RAG&\underline{6.22}&6.51&2.98&5.02&5.03&2.13&	&6.17&6.34&3.39&5.25&5.58&3.60\\
     &APTNESS&\textbf{6.28}&\textbf{6.68}&3.37&\textbf{5.23}&5.23&2.28&	&6.44&6.41&3.98&5.48&5.93&4.39\\\midrule
    \multicolumn{2}{r}{Pearson Correlation with Emp.} &-&0.27&-0.60&0.92$\uparrow\uparrow$&0.62$\uparrow$&-0.60&	&-&0.37&0.19&0.97$\uparrow\uparrow$&0.73$\uparrow$&0.24\\
    \bottomrule
    \end{tabular}
    \label{tab:comparison}
\end{table*}

\subsection{Main Results}

\paragraph{Comparison of APTNESS against different baselines across different foundation models.}
We evaluate each of the models discussed above using the two collected datasets. 
\begin{itemize}
    \item \textbf{GEN}. This method tests the conversational ability of the LLM by allowing it to directly generate a response $G_1$ to the dialogue history $C$.
    \item \textbf{RAG}. This method enhances the LLM with retrieval augmentation. Using the APT database, an index is created which allows the LLM to utilize the retrieved responses along with $G_1$ to generate a retrieval-augmented response.
    \item \textbf{APTNESS}. This method generates responses using our proposed APTNESS framework. Integrate retrieval augmentation with the strategy incorporation module, utilizing $G_1$, the retrieved responses $R_1, R_2,...,R_{K}$, and the predicted strategy $S_1,S_2,...,S_{K}$ and their definitions to jointly create responses.
\end{itemize}

Table~\ref{tab:comparison} displays the experimental results on ED and ET. From these results, on the primary evaluation metric, empathy, we observe that: 
(1) Llama3-8B+APTNESS achieves a score of 6.28 in ED and Llama2-7B+APTNESS reaches 6.50 in ET. Both are the state-of-the-art results achieved by the APTNESS method, corroborating the effectiveness of our approach.
(2) Compared to Gen, the RAG method significantly outperforms in empathy, with Llama2-7B-based scores being higher by 0.52 and 0.39 in ED and ET respectively, and Llama3-8B-based scores being higher by 0.50 and 0.27 respectively. This validates the effectiveness of our constructed empathy database, APT, in enhancing the empathetic capabilities of LLMs. 
(3) We find that Llama3-8B performs better in empathy on ED, scoring 0.06 higher than Llama2-7B, but on ET, Llama2-7B scores 0.06 higher than Llama3-8B. We speculate that this is due to some degradation in the generative capabilities of Llama3-8B in longer dialogues.
\paragraph{Correlated Features with Empathy.} In our research, we employ the Pearson coefficient to ascertain the correlation between various sub-metrics and the primary evaluation metric \textit{Empathy}. 
The Pearson coefficient serves as a quantifiable measure of the linear correlation existing between two distinct data sets. It is calculated as the ratio of the covariance between the two variables to the product of their respective standard deviations~\cite{cohen2009pearson}.
We find that the sub-metric identification has correlations of 0.92 and 0.97 with the empathy on ED and ET respectively, indicating a very strong correlation. The comforting sub-metric has correlations of 0.62 and 0.73 respectively, indicating a strong correlation. 

We believe these two metrics respectively represent the capability to identify emotions and to comfort others. This observation is in harmony with our classification of empathy into two key components: \textit{Cognitive Empathy} and \textit{Affective Empathy}.
Our study reveals:
(1) Compared to GEN, RAG significantly improves the model's empathy capabilities in both affective and cognitive empathy. We attribute this improvement to our constructed APT database, which contains rich empathetic dialogue materials that help the model better understand the emotional context and generate more nuanced expressions. However, we observe a slight decline in suggestion and informativity when RAG is introduced. We speculate that this may be due to RAG encouraging the LLM to reduce preachy responses and incorporate a wider variety of expressions. 
(2) Additionally, we observe that APTNESS can further enhance the empathetic abilities of LLMs beyond the improvements seen with RAG alone. We believe that incorporating emotional support strategies introduces additional dialogue techniques, enabling the model to better address issues and provide comforting responses. This finding supports our hypothesis that enhancing empathy requires improvements in both affective and cognitive empathy. With the addition of strategies, suggestion and informativity show some improvement, indicating that the introduction of emotional support strategies can address issues of insufficient expression.


\begin{table*}[t] 
    \centering
    \caption{An example of ED and the responses generated responses from GEN, RAG and APTNESS over Llama2-7B. \textcolor{orange}{The orange contents} show the effect of improvement methods.} 
    \begin{tabular}{cp{15.8cm}}
    \toprule
     \textbf{Speaker:}& One of my neighbors (next building and two floors up) plays talk radio every morning Very Loudly.\\
     \textbf{Listener:} &Urgh that sounds like the worst! How early?\\
     \textbf{Speaker:} &Not until about 8am, but it's so loud I can hear it in my first floor apartment with the doors and windows closed. Idiot man!\\
     \midrule
     \multirow{6}{*}{GEN}&I understand your frustration. It can be very disruptive and inconsiderate when someone plays their music too loudly, especially when it's disturbing your peace and quiet in your own home. While it's understandable that you might want to confront your neighbor about the issue, it's important to approach the situation in a calm and respectful manner. \textcolor{orange}{Here are some steps you could take:
1. Talk to your neighbor: ...
2. Use earplugs or earmuffs: ...
3. Contact your landlord or} \textcolor{orange}{building management:...} 
Remember, approaching the situation in a calm and respectful manner is key. It's important to be considerate of your neighbor's feelings and to find a solution that works for everyone involved.\\
     \midrule
     \multirow{4}{*}{RAG}&\textcolor{orange}{I can understand why you would find your neighbor's loud talk radio annoying}, especially when it's disrupting your peace and quiet in the morning. It's important to communicate your concerns with your neighbor in a respectful manner, and you may want to consider using earplugs or white noise machines to help block out the sound. If the issue persists, you could also consider contacting building management or the authorities for assistance. Remember, communication and respect are key to resolving issues with neighbors.\\

     \midrule
     \multirow{8}{*}{APTNESS}  &   \textcolor{orange}{Oh no, that's really frustrating to deal with a neighbor who's consistently playing loud talk radio. It's understandable that} \textcolor{orange}{you're feeling annoyed and disrespected,} especially since you're trying to enjoy some peace and quiet in your own home. It's important to remember that approaching the situation in a respectful and constructive manner is key to finding a resolution that works for everyone involved. Have you considered talking to your neighbor directly and expressing your concerns in a polite and respectful way? Sometimes, people may not be aware of how much noise they're making, and a direct conversation can help them understand your perspective. If that doesn't work, you could also consider contacting your landlord or building management for assistance. \textcolor{orange}{Remember, you have the right to enjoy your own home without} \textcolor{orange}{excessive noise disturbance.} Don't hesitate to seek help if you need it, and try to approach the situation with empathy and understanding.\\
    \bottomrule
    \end{tabular}
    \label{tab:case study}
\end{table*}


\subsection{Effectiveness of Different Strategy Schemes}

To investigate the impact of different strategy annotation schemes on performance, we separately train a LoRA module for Llama3-8B using two empathetic dialogue datasets, ESConv and ExTES, each with different strategy annotation schemes. Specifically, ExTES uses more strategies than ESConv, but each response in ExTES has only one strategy annotated. In contrast, a response in ESConv may have multiple strategies annotated. The experimental results are listed in Table~\ref{tab:strategy}.

\begin{table}[!htp] 
    \centering
    \caption{Comparison of using different strategy schemes over Llama3-8B. The maximum value for each evaluation metric is bolded.}
    \begin{tabular}{cccccccc}
    \toprule
     Data&Strategy&Emp.&	Coh.	&Inf.&	Iden.&	Comf.&	Sug. \\

     \midrule
     \multirow{2}{*}{ED}&ESConv&6.23&6.61&\textbf{3.63}&5.07&5.09&\textbf{3.06}\\
     &ExTES&\textbf{6.28}&\textbf{6.68}&3.37&\textbf{5.23}&\textbf{5.23}&2.28\\
     \midrule
     \multirow{2}{*}{ET}&ESConv&6.22&\textbf{6.52}&4.12&5.46&5.58&4.25\\
     &ExTES&\textbf{6.44}&6.41&3.98&\textbf{5.48}&\textbf{5.93}&\textbf{4.39}\\
    \bottomrule
    \end{tabular}
    \label{tab:strategy}
\end{table}

We observe that when the strategy injection module trained with ExTES is used for the APTNESS method, there is an improvement of 0.05, 0.16, and 0.14 on ED for Empathy, Identification, and Comforting, respectively, compared to the module trained with ESConv. On ET, the improvements are 0.22, 0.02, and 0.14 points for Empathy, Identification, and Comforting, respectively. Considering the overall experimental results, we believe that fine-grained strategy annotations are more effective than multiple strategy annotations for integrating emotional support strategies. We posit that finer-grained strategies provide the LLM with more nuanced guidance, enhancing both affective and cognitive empathy, and thereby significantly improving the model's empathetic capabilities.

\subsection{Necessity of the Strategy Integration} 

 In this section, we validate the necessity of the emotional support strategy injection module. We conduct experiments using ChatGPT as the foundation model with the two datasets mentioned above. In these experiments, strategy predictions are generated by ChatGPT itself, rather than using a fine-tuned strategy integration module. The results are shown in Table~\ref{tab:chatgpt}.

\begin{table}[!htp] 
    \centering
    \caption{Performance of ChatGPT with APTNESS . The maximum value for each evaluation metric is bolded.}
    \begin{tabular}{cccccccc}
    \toprule
     Data&Strategy&Emp.&	Coh.	&Inf.&	Iden.&	Comf.&	Sug. \\

     \midrule
     \multirow{3}{*}{ED}&GEN&5.20&6.40&3.33&4.15&4.48&2.90\\
     &RAG&5.94&6.68&3.38&4.72&5.07&3.12\\
     &APTNESS&\textbf{6.19}&\textbf{6.73}&\textbf{3.71}&\textbf{4.93}&\textbf{5.20}&\textbf{3.51}\\
     \midrule
     \multirow{3}{*}{ET}&GEN&5.43&6.68&3.64&4.55&4.93&3.99\\
     &RAG&\textbf{6.08}&\textbf{6.78}&4.23&5.11&\textbf{5.73}&4.73\\
     &APTNESS&5.85&6.62&\textbf{4.35}&\textbf{5.13}&5.39&\textbf{4.84}\\
    \bottomrule
    \end{tabular}
    \label{tab:chatgpt}
\end{table}

The results show that on the ED dataset, ChatGPT combined with APTNESS significantly improves empathetic ability, achieving an empathy score increase of 0.25 over RAG. However, on the ET dataset, its performance declines, with empathy and comforting scores decreasing by 0.23 and 0.34, respectively, compared to RAG. We believe this is because, although ChatGPT has the ability to generate strategies, its performance is not stable. This instability may lead to inaccurate responses that negatively impact the LLM's empathetic responses, particularly in comforting. As a result, the LLM's cognitive empathy ability is weakened, ultimately leading to a decrease in overall empathetic capability.
\subsection{Case Study}
Table~\ref{tab:case study} lists the responses from the three methods discussed. GEN can discern emotional issues to some extent but offers only mechanical suggestions, resulting in weak comforting and poor cognitive performance. RAG improves GEN's monotony by providing more fluent and appropriate expressions and the result shows that the LLM no longer lists opinions mechanically like GEN. What's more, it begins to empathically identify and comforting with the dialogue's context as the orange contents.
Our proposed framework, APTNESS, fully comprehends the user's feelings, identifies issues more accurately, and offers better support. 
The phrase "Remember, you..." in orange demonstrates APTNESS's superior understanding of the speaker’s situation and steadfast emotional support. It enhances empathetic abilities, indicating that the model is more willing to put itself in the speaker's shoes, feeling the speaker's emotions and situation, and comforting the user from these perspectives.

\section{Conclusion}
In this paper, we present APTNESS, an innovative framework grounded in appraisal theory and emotional support strategies. Our research includes the creation of the APT database, an empathetic response database, and a two-stage empathetic response generation method.
\begin{acks}
This work was partially supported by China Postdoctoral Science Foundation (2023M733654),  Guangdong Basic and Applied Basic Research Foundation (2023A1515110496).
\end{acks}
\bibliographystyle{ACM-Reference-Format}
\bibliography{anthology.aa,anthology.ab,custom}


\begin{thebibliography}{34}


\ifx \showCODEN    \undefined \def \showCODEN     #1{\unskip}     \fi
\ifx \showDOI      \undefined \def \showDOI       #1{#1}\fi
\ifx \showISBNx    \undefined \def \showISBNx     #1{\unskip}     \fi
\ifx \showISBNxiii \undefined \def \showISBNxiii  #1{\unskip}     \fi
\ifx \showISSN     \undefined \def \showISSN      #1{\unskip}     \fi
\ifx \showLCCN     \undefined \def \showLCCN      #1{\unskip}     \fi
\ifx \shownote     \undefined \def \shownote      #1{#1}          \fi
\ifx \showarticletitle \undefined \def \showarticletitle #1{#1}   \fi
\ifx \showURL      \undefined \def \showURL       {\relax}        \fi
\providecommand\bibfield[2]{#2}
\providecommand\bibinfo[2]{#2}
\providecommand\natexlab[1]{#1}
\providecommand\showeprint[2][]{arXiv:#2}

\bibitem[Cohen et~al\mbox{.}(2009)]%
        {cohen2009pearson}
\bibfield{author}{\bibinfo{person}{Israel Cohen}, \bibinfo{person}{Yiteng Huang}, \bibinfo{person}{Jingdong Chen}, \bibinfo{person}{Jacob Benesty}, \bibinfo{person}{Jacob Benesty}, \bibinfo{person}{Jingdong Chen}, \bibinfo{person}{Yiteng Huang}, {and} \bibinfo{person}{Israel Cohen}.} \bibinfo{year}{2009}\natexlab{}.
\newblock \showarticletitle{Pearson correlation coefficient}.
\newblock \bibinfo{journal}{\emph{Noise reduction in speech processing}} (\bibinfo{year}{2009}), \bibinfo{pages}{1--4}.
\newblock


\bibitem[Ekman(1993)]%
        {ekman1993facial}
\bibfield{author}{\bibinfo{person}{Paul Ekman}.} \bibinfo{year}{1993}\natexlab{}.
\newblock \showarticletitle{Facial expression and emotion.}
\newblock \bibinfo{journal}{\emph{American psychologist}} \bibinfo{volume}{48}, \bibinfo{number}{4} (\bibinfo{year}{1993}), \bibinfo{pages}{384}.
\newblock


\bibitem[Gao et~al\mbox{.}(2023)]%
        {gao2023retrieval}
\bibfield{author}{\bibinfo{person}{Yunfan Gao}, \bibinfo{person}{Yun Xiong}, \bibinfo{person}{Xinyu Gao}, \bibinfo{person}{Kangxiang Jia}, \bibinfo{person}{Jinliu Pan}, \bibinfo{person}{Yuxi Bi}, \bibinfo{person}{Yi Dai}, \bibinfo{person}{Jiawei Sun}, {and} \bibinfo{person}{Haofen Wang}.} \bibinfo{year}{2023}\natexlab{}.
\newblock \showarticletitle{Retrieval-augmented generation for large language models: A survey}.
\newblock \bibinfo{journal}{\emph{arXiv preprint arXiv:2312.10997}} (\bibinfo{year}{2023}).
\newblock


\bibitem[Hu et~al\mbox{.}(2021)]%
        {hu2021lora}
\bibfield{author}{\bibinfo{person}{Edward~J Hu}, \bibinfo{person}{Yelong Shen}, \bibinfo{person}{Phillip Wallis}, \bibinfo{person}{Zeyuan Allen-Zhu}, \bibinfo{person}{Yuanzhi Li}, \bibinfo{person}{Shean Wang}, \bibinfo{person}{Lu Wang}, {and} \bibinfo{person}{Weizhu Chen}.} \bibinfo{year}{2021}\natexlab{}.
\newblock \showarticletitle{Lora: Low-rank adaptation of large language models}.
\newblock \bibinfo{journal}{\emph{arXiv preprint arXiv:2106.09685}} (\bibinfo{year}{2021}).
\newblock


\bibitem[Huang et~al\mbox{.}(2023)]%
        {huang2023emotionally}
\bibfield{author}{\bibinfo{person}{Jen{-}tse Huang}, \bibinfo{person}{Man~Ho Lam}, \bibinfo{person}{Eric~John Li}, \bibinfo{person}{Shujie Ren}, \bibinfo{person}{Wenxuan Wang}, \bibinfo{person}{Wenxiang Jiao}, \bibinfo{person}{Zhaopeng Tu}, {and} \bibinfo{person}{Michael~R. Lyu}.} \bibinfo{year}{2023}\natexlab{}.
\newblock \showarticletitle{Emotionally Numb or Empathetic? Evaluating How {LLM}s Feel Using Emotion{B}ench}.
\newblock \bibinfo{journal}{\emph{arXiv preprint arXiv:2308.03656}} (\bibinfo{year}{2023}).
\newblock


\bibitem[Hwang et~al\mbox{.}(2021)]%
        {hwang2021comet}
\bibfield{author}{\bibinfo{person}{Jena~D Hwang}, \bibinfo{person}{Chandra Bhagavatula}, \bibinfo{person}{Ronan Le~Bras}, \bibinfo{person}{Jeff Da}, \bibinfo{person}{Keisuke Sakaguchi}, \bibinfo{person}{Antoine Bosselut}, {and} \bibinfo{person}{Yejin Choi}.} \bibinfo{year}{2021}\natexlab{}.
\newblock \showarticletitle{(comet-) atomic 2020: On symbolic and neural commonsense knowledge graphs}. In \bibinfo{booktitle}{\emph{Proceedings of the AAAI Conference on Artificial Intelligence}}, Vol.~\bibinfo{volume}{35}. \bibinfo{pages}{6384--6392}.
\newblock


\bibitem[Izacard et~al\mbox{.}(2021)]%
        {izacard2021unsupervised}
\bibfield{author}{\bibinfo{person}{Gautier Izacard}, \bibinfo{person}{Mathilde Caron}, \bibinfo{person}{Lucas Hosseini}, \bibinfo{person}{Sebastian Riedel}, \bibinfo{person}{Piotr Bojanowski}, \bibinfo{person}{Armand Joulin}, {and} \bibinfo{person}{Edouard Grave}.} \bibinfo{year}{2021}\natexlab{}.
\newblock \showarticletitle{Unsupervised dense information retrieval with contrastive learning}.
\newblock \bibinfo{journal}{\emph{arXiv preprint arXiv:2112.09118}} (\bibinfo{year}{2021}).
\newblock


\bibitem[Izacard and Grave(2021)]%
        {izacard-grave-2021-leveraging}
\bibfield{author}{\bibinfo{person}{Gautier Izacard} {and} \bibinfo{person}{Edouard Grave}.} \bibinfo{year}{2021}\natexlab{}.
\newblock \showarticletitle{Leveraging Passage Retrieval with Generative Models for Open Domain Question Answering}. In \bibinfo{booktitle}{\emph{Proceedings of the 16th Conference of the European Chapter of the Association for Computational Linguistics: Main Volume}}, \bibfield{editor}{\bibinfo{person}{Paola Merlo}, \bibinfo{person}{Jorg Tiedemann}, {and} \bibinfo{person}{Reut Tsarfaty}} (Eds.). \bibinfo{publisher}{Association for Computational Linguistics}, \bibinfo{address}{Online}, \bibinfo{pages}{874--880}.
\newblock
\urldef\tempurl%
\url{https://doi.org/10.18653/v1/2021.eacl-main.74}
\showDOI{\tempurl}


\bibitem[Karpukhin et~al\mbox{.}(2020)]%
        {karpukhin-etal-2020-dense}
\bibfield{author}{\bibinfo{person}{Vladimir Karpukhin}, \bibinfo{person}{Barlas Oguz}, \bibinfo{person}{Sewon Min}, \bibinfo{person}{Patrick Lewis}, \bibinfo{person}{Ledell Wu}, \bibinfo{person}{Sergey Edunov}, \bibinfo{person}{Danqi Chen}, {and} \bibinfo{person}{Wen-tau Yih}.} \bibinfo{year}{2020}\natexlab{}.
\newblock \showarticletitle{Dense Passage Retrieval for Open-Domain Question Answering}. In \bibinfo{booktitle}{\emph{Proceedings of the 2020 Conference on Empirical Methods in Natural Language Processing (EMNLP)}}, \bibfield{editor}{\bibinfo{person}{Bonnie Webber}, \bibinfo{person}{Trevor Cohn}, \bibinfo{person}{Yulan He}, {and} \bibinfo{person}{Yang Liu}} (Eds.). \bibinfo{publisher}{Association for Computational Linguistics}, \bibinfo{address}{Online}, \bibinfo{pages}{6769--6781}.
\newblock
\urldef\tempurl%
\url{https://doi.org/10.18653/v1/2020.emnlp-main.550}
\showDOI{\tempurl}


\bibitem[Keskin(2014)]%
        {keskin2014isn}
\bibfield{author}{\bibinfo{person}{Sevgi~Co{\c{s}}kun Keskin}.} \bibinfo{year}{2014}\natexlab{}.
\newblock \showarticletitle{From what isn’t empathy to empathic learning process}.
\newblock \bibinfo{journal}{\emph{Procedia-Social and Behavioral Sciences}}  \bibinfo{volume}{116} (\bibinfo{year}{2014}), \bibinfo{pages}{4932--4938}.
\newblock


\bibitem[Kim et~al\mbox{.}(2023)]%
        {kim-etal-2023-tree}
\bibfield{author}{\bibinfo{person}{Gangwoo Kim}, \bibinfo{person}{Sungdong Kim}, \bibinfo{person}{Byeongguk Jeon}, \bibinfo{person}{Joonsuk Park}, {and} \bibinfo{person}{Jaewoo Kang}.} \bibinfo{year}{2023}\natexlab{}.
\newblock \showarticletitle{Tree of Clarifications: Answering Ambiguous Questions with Retrieval-Augmented Large Language Models}. In \bibinfo{booktitle}{\emph{Proceedings of the 2023 Conference on Empirical Methods in Natural Language Processing}}, \bibfield{editor}{\bibinfo{person}{Houda Bouamor}, \bibinfo{person}{Juan Pino}, {and} \bibinfo{person}{Kalika Bali}} (Eds.). \bibinfo{publisher}{Association for Computational Linguistics}, \bibinfo{address}{Singapore}, \bibinfo{pages}{996--1009}.
\newblock
\urldef\tempurl%
\url{https://doi.org/10.18653/v1/2023.emnlp-main.63}
\showDOI{\tempurl}


\bibitem[Komorita and Graham(1965)]%
        {komorita1965number}
\bibfield{author}{\bibinfo{person}{Samuel~S Komorita} {and} \bibinfo{person}{William~K Graham}.} \bibinfo{year}{1965}\natexlab{}.
\newblock \showarticletitle{Number of scale points and the reliability of scales}.
\newblock \bibinfo{journal}{\emph{Educational and Psychological Measurement}} \bibinfo{volume}{25}, \bibinfo{number}{4} (\bibinfo{year}{1965}), \bibinfo{pages}{987--995}.
\newblock


\bibitem[Lawrence et~al\mbox{.}(2004)]%
        {lawrence2004measuring}
\bibfield{author}{\bibinfo{person}{Emma~J Lawrence}, \bibinfo{person}{Philip Shaw}, \bibinfo{person}{Dawn Baker}, \bibinfo{person}{Simon Baron-Cohen}, {and} \bibinfo{person}{Anthony~S David}.} \bibinfo{year}{2004}\natexlab{}.
\newblock \showarticletitle{Measuring empathy: reliability and validity of the Empathy Quotient}.
\newblock \bibinfo{journal}{\emph{Psychological medicine}} \bibinfo{volume}{34}, \bibinfo{number}{5} (\bibinfo{year}{2004}), \bibinfo{pages}{911--920}.
\newblock


\bibitem[Liu et~al\mbox{.}(2021)]%
        {liu-etal-2021-towards}
\bibfield{author}{\bibinfo{person}{Siyang Liu}, \bibinfo{person}{Chujie Zheng}, \bibinfo{person}{Orianna Demasi}, \bibinfo{person}{Sahand Sabour}, \bibinfo{person}{Yu Li}, \bibinfo{person}{Zhou Yu}, \bibinfo{person}{Yong Jiang}, {and} \bibinfo{person}{Minlie Huang}.} \bibinfo{year}{2021}\natexlab{}.
\newblock \showarticletitle{Towards Emotional Support Dialog Systems}. In \bibinfo{booktitle}{\emph{Proceedings of the 59th Annual Meeting of the Association for Computational Linguistics and the 11th International Joint Conference on Natural Language Processing (Volume 1: Long Papers)}}, \bibfield{editor}{\bibinfo{person}{Chengqing Zong}, \bibinfo{person}{Fei Xia}, \bibinfo{person}{Wenjie Li}, {and} \bibinfo{person}{Roberto Navigli}} (Eds.). \bibinfo{publisher}{Association for Computational Linguistics}, \bibinfo{address}{Online}, \bibinfo{pages}{3469--3483}.
\newblock
\urldef\tempurl%
\url{https://doi.org/10.18653/v1/2021.acl-long.269}
\showDOI{\tempurl}


\bibitem[Ma et~al\mbox{.}(2023)]%
        {ma2023query}
\bibfield{author}{\bibinfo{person}{Xinbei Ma}, \bibinfo{person}{Yeyun Gong}, \bibinfo{person}{Pengcheng He}, \bibinfo{person}{Hai Zhao}, {and} \bibinfo{person}{Nan Duan}.} \bibinfo{year}{2023}\natexlab{}.
\newblock \showarticletitle{Query rewriting for retrieval-augmented large language models}.
\newblock \bibinfo{journal}{\emph{arXiv preprint arXiv:2305.14283}} (\bibinfo{year}{2023}).
\newblock


\bibitem[Mackie et~al\mbox{.}(2023)]%
        {mackie2023generative}
\bibfield{author}{\bibinfo{person}{Iain Mackie}, \bibinfo{person}{Shubham Chatterjee}, {and} \bibinfo{person}{Jeffrey Dalton}.} \bibinfo{year}{2023}\natexlab{}.
\newblock \showarticletitle{Generative relevance feedback with large language models}. In \bibinfo{booktitle}{\emph{Proceedings of the 46th International ACM SIGIR Conference on Research and Development in Information Retrieval}}. \bibinfo{pages}{2026--2031}.
\newblock


\bibitem[Nussbaum et~al\mbox{.}(2024)]%
        {nussbaum2024nomic}
\bibfield{author}{\bibinfo{person}{Zach Nussbaum}, \bibinfo{person}{John~X Morris}, \bibinfo{person}{Brandon Duderstadt}, {and} \bibinfo{person}{Andriy Mulyar}.} \bibinfo{year}{2024}\natexlab{}.
\newblock \showarticletitle{Nomic Embed: Training a Reproducible Long Context Text Embedder}.
\newblock \bibinfo{journal}{\emph{arXiv preprint arXiv:2402.01613}} (\bibinfo{year}{2024}).
\newblock


\bibitem[Peng et~al\mbox{.}(2022)]%
        {peng2022control}
\bibfield{author}{\bibinfo{person}{Wei Peng}, \bibinfo{person}{Yue Hu}, \bibinfo{person}{Luxi Xing}, \bibinfo{person}{Yuqiang Xie}, \bibinfo{person}{Yajing Sun}, {and} \bibinfo{person}{Yunpeng Li}.} \bibinfo{year}{2022}\natexlab{}.
\newblock \showarticletitle{Control globally, understand locally: A global-to-local hierarchical graph network for emotional support conversation}.
\newblock \bibinfo{journal}{\emph{arXiv preprint arXiv:2204.12749}} (\bibinfo{year}{2022}).
\newblock


\bibitem[Qian et~al\mbox{.}(2023)]%
        {qian-etal-2023-harnessing}
\bibfield{author}{\bibinfo{person}{Yushan Qian}, \bibinfo{person}{Weinan Zhang}, {and} \bibinfo{person}{Ting Liu}.} \bibinfo{year}{2023}\natexlab{}.
\newblock \showarticletitle{Harnessing the Power of Large Language Models for Empathetic Response Generation: Empirical Investigations and Improvements}. In \bibinfo{booktitle}{\emph{Findings of the Association for Computational Linguistics: EMNLP 2023}}, \bibfield{editor}{\bibinfo{person}{Houda Bouamor}, \bibinfo{person}{Juan Pino}, {and} \bibinfo{person}{Kalika Bali}} (Eds.). \bibinfo{publisher}{Association for Computational Linguistics}, \bibinfo{address}{Singapore}, \bibinfo{pages}{6516--6528}.
\newblock
\urldef\tempurl%
\url{https://doi.org/10.18653/v1/2023.findings-emnlp.433}
\showDOI{\tempurl}


\bibitem[Rashkin et~al\mbox{.}(2019)]%
        {rashkin-etal-2019-towards}
\bibfield{author}{\bibinfo{person}{Hannah Rashkin}, \bibinfo{person}{Eric~Michael Smith}, \bibinfo{person}{Margaret Li}, {and} \bibinfo{person}{Y-Lan Boureau}.} \bibinfo{year}{2019}\natexlab{}.
\newblock \showarticletitle{Towards Empathetic Open-domain Conversation Models: A New Benchmark and Dataset}. In \bibinfo{booktitle}{\emph{Proceedings of the 57th Annual Meeting of the Association for Computational Linguistics}}, \bibfield{editor}{\bibinfo{person}{Anna Korhonen}, \bibinfo{person}{David Traum}, {and} \bibinfo{person}{Llu{\'\i}s M{\`a}rquez}} (Eds.). \bibinfo{publisher}{Association for Computational Linguistics}, \bibinfo{address}{Florence, Italy}, \bibinfo{pages}{5370--5381}.
\newblock
\urldef\tempurl%
\url{https://doi.org/10.18653/v1/P19-1534}
\showDOI{\tempurl}


\bibitem[Reniers et~al\mbox{.}(2011)]%
        {reniers2011qcae}
\bibfield{author}{\bibinfo{person}{Renate~LEP Reniers}, \bibinfo{person}{Rhiannon Corcoran}, \bibinfo{person}{Richard Drake}, \bibinfo{person}{Nick~M Shryane}, {and} \bibinfo{person}{Birgit~A V{\"o}llm}.} \bibinfo{year}{2011}\natexlab{}.
\newblock \showarticletitle{The QCAE: A questionnaire of cognitive and affective empathy}.
\newblock \bibinfo{journal}{\emph{Journal of personality assessment}} \bibinfo{volume}{93}, \bibinfo{number}{1} (\bibinfo{year}{2011}), \bibinfo{pages}{84--95}.
\newblock


\bibitem[Sabour et~al\mbox{.}(2022)]%
        {sabour2022cem}
\bibfield{author}{\bibinfo{person}{Sahand Sabour}, \bibinfo{person}{Chujie Zheng}, {and} \bibinfo{person}{Minlie Huang}.} \bibinfo{year}{2022}\natexlab{}.
\newblock \showarticletitle{Cem: Commonsense-aware empathetic response generation}. In \bibinfo{booktitle}{\emph{Proceedings of the AAAI Conference on Artificial Intelligence}}, Vol.~\bibinfo{volume}{36}. \bibinfo{pages}{11229--11237}.
\newblock


\bibitem[Smith(2006)]%
        {smith2006cognitive}
\bibfield{author}{\bibinfo{person}{Adam Smith}.} \bibinfo{year}{2006}\natexlab{}.
\newblock \showarticletitle{Cognitive empathy and emotional empathy in human behavior and evolution}.
\newblock \bibinfo{journal}{\emph{The Psychological Record}} \bibinfo{volume}{56}, \bibinfo{number}{1} (\bibinfo{year}{2006}), \bibinfo{pages}{3--21}.
\newblock


\bibitem[Speer et~al\mbox{.}(2017)]%
        {speer2017conceptnet}
\bibfield{author}{\bibinfo{person}{Robyn Speer}, \bibinfo{person}{Joshua Chin}, {and} \bibinfo{person}{Catherine Havasi}.} \bibinfo{year}{2017}\natexlab{}.
\newblock \showarticletitle{Conceptnet 5.5: An open multilingual graph of general knowledge}. In \bibinfo{booktitle}{\emph{Proceedings of the AAAI conference on artificial intelligence}}, Vol.~\bibinfo{volume}{31}.
\newblock


\bibitem[Sun et~al\mbox{.}(2024)]%
        {sun2024rational}
\bibfield{author}{\bibinfo{person}{Linzhuang Sun}, \bibinfo{person}{Nan Xu}, \bibinfo{person}{Jingxuan Wei}, \bibinfo{person}{Bihui Yu}, \bibinfo{person}{Liping Bu}, {and} \bibinfo{person}{Yin Luo}.} \bibinfo{year}{2024}\natexlab{}.
\newblock \bibinfo{title}{Rational Sensibility: LLM Enhanced Empathetic Response Generation Guided by Self-presentation Theory}.
\newblock
\newblock
\showeprint[arxiv]{2312.08702}~[cs.AI]


\bibitem[Touvron et~al\mbox{.}(2023)]%
        {touvron2023llama}
\bibfield{author}{\bibinfo{person}{Hugo Touvron}, \bibinfo{person}{Thibaut Lavril}, \bibinfo{person}{Gautier Izacard}, \bibinfo{person}{Xavier Martinet}, \bibinfo{person}{Marie-Anne Lachaux}, \bibinfo{person}{Timoth{\'e}e Lacroix}, \bibinfo{person}{Baptiste Rozi{\`e}re}, \bibinfo{person}{Naman Goyal}, \bibinfo{person}{Eric Hambro}, \bibinfo{person}{Faisal Azhar}, {et~al\mbox{.}}} \bibinfo{year}{2023}\natexlab{}.
\newblock \showarticletitle{Llama: Open and efficient foundation language models}.
\newblock \bibinfo{journal}{\emph{arXiv preprint arXiv:2302.13971}} (\bibinfo{year}{2023}).
\newblock


\bibitem[Tu et~al\mbox{.}(2022)]%
        {tu-etal-2022-misc}
\bibfield{author}{\bibinfo{person}{Quan Tu}, \bibinfo{person}{Yanran Li}, \bibinfo{person}{Jianwei Cui}, \bibinfo{person}{Bin Wang}, \bibinfo{person}{Ji-Rong Wen}, {and} \bibinfo{person}{Rui Yan}.} \bibinfo{year}{2022}\natexlab{}.
\newblock \showarticletitle{{MISC}: A Mixed Strategy-Aware Model integrating {COMET} for Emotional Support Conversation}. In \bibinfo{booktitle}{\emph{Proceedings of the 60th Annual Meeting of the Association for Computational Linguistics (Volume 1: Long Papers)}}, \bibfield{editor}{\bibinfo{person}{Smaranda Muresan}, \bibinfo{person}{Preslav Nakov}, {and} \bibinfo{person}{Aline Villavicencio}} (Eds.). \bibinfo{publisher}{Association for Computational Linguistics}, \bibinfo{address}{Dublin, Ireland}, \bibinfo{pages}{308--319}.
\newblock
\urldef\tempurl%
\url{https://doi.org/10.18653/v1/2022.acl-long.25}
\showDOI{\tempurl}


\bibitem[Wang et~al\mbox{.}(2023)]%
        {wang2023query2doc}
\bibfield{author}{\bibinfo{person}{Liang Wang}, \bibinfo{person}{Nan Yang}, {and} \bibinfo{person}{Furu Wei}.} \bibinfo{year}{2023}\natexlab{}.
\newblock \showarticletitle{Query2doc: Query expansion with large language models}.
\newblock \bibinfo{journal}{\emph{arXiv preprint arXiv:2303.07678}} (\bibinfo{year}{2023}).
\newblock


\bibitem[Xu et~al\mbox{.}(2008)]%
        {Xu2007Constructing}
\bibfield{author}{\bibinfo{person}{L. Xu}, \bibinfo{person}{Hongfei Lin}, \bibinfo{person}{Y. Pan}, \bibinfo{person}{H. Ren}, {and} \bibinfo{person}{J. Chen}.} \bibinfo{year}{2008}\natexlab{}.
\newblock \showarticletitle{Constructing the affective lexicon ontology}.
\newblock \bibinfo{journal}{\emph{Journal of the China Society for Scientific and Technical Information}}  \bibinfo{volume}{27} (\bibinfo{date}{01} \bibinfo{year}{2008}), \bibinfo{pages}{180--185}.
\newblock


\bibitem[Yu et~al\mbox{.}(2022)]%
        {yu2022generate}
\bibfield{author}{\bibinfo{person}{Wenhao Yu}, \bibinfo{person}{Dan Iter}, \bibinfo{person}{Shuohang Wang}, \bibinfo{person}{Yichong Xu}, \bibinfo{person}{Mingxuan Ju}, \bibinfo{person}{Soumya Sanyal}, \bibinfo{person}{Chenguang Zhu}, \bibinfo{person}{Michael Zeng}, {and} \bibinfo{person}{Meng Jiang}.} \bibinfo{year}{2022}\natexlab{}.
\newblock \showarticletitle{Generate rather than retrieve: Large language models are strong context generators}.
\newblock \bibinfo{journal}{\emph{arXiv preprint arXiv:2209.10063}} (\bibinfo{year}{2022}).
\newblock


\bibitem[Zhang et~al\mbox{.}(2023)]%
        {zhang2023merging}
\bibfield{author}{\bibinfo{person}{Yunxiang Zhang}, \bibinfo{person}{Muhammad Khalifa}, \bibinfo{person}{Lajanugen Logeswaran}, \bibinfo{person}{Moontae Lee}, \bibinfo{person}{Honglak Lee}, {and} \bibinfo{person}{Lu Wang}.} \bibinfo{year}{2023}\natexlab{}.
\newblock \showarticletitle{Merging generated and retrieved knowledge for open-domain QA}.
\newblock \bibinfo{journal}{\emph{arXiv preprint arXiv:2310.14393}} (\bibinfo{year}{2023}).
\newblock


\bibitem[Zheng et~al\mbox{.}(2023)]%
        {zheng2023building}
\bibfield{author}{\bibinfo{person}{Zhonghua Zheng}, \bibinfo{person}{Lizi Liao}, \bibinfo{person}{Yang Deng}, {and} \bibinfo{person}{Liqiang Nie}.} \bibinfo{year}{2023}\natexlab{}.
\newblock \showarticletitle{Building emotional support chatbots in the era of llms}.
\newblock \bibinfo{journal}{\emph{arXiv preprint arXiv:2308.11584}} (\bibinfo{year}{2023}).
\newblock


\bibitem[Zhou et~al\mbox{.}(2023a)]%
        {zhou-etal-2023-facilitating}
\bibfield{author}{\bibinfo{person}{Jinfeng Zhou}, \bibinfo{person}{Zhuang Chen}, \bibinfo{person}{Bo Wang}, {and} \bibinfo{person}{Minlie Huang}.} \bibinfo{year}{2023}\natexlab{a}.
\newblock \showarticletitle{Facilitating Multi-turn Emotional Support Conversation with Positive Emotion Elicitation: A Reinforcement Learning Approach}. In \bibinfo{booktitle}{\emph{Proceedings of the 61st Annual Meeting of the Association for Computational Linguistics (Volume 1: Long Papers)}}, \bibfield{editor}{\bibinfo{person}{Anna Rogers}, \bibinfo{person}{Jordan Boyd-Graber}, {and} \bibinfo{person}{Naoaki Okazaki}} (Eds.). \bibinfo{publisher}{Association for Computational Linguistics}, \bibinfo{address}{Toronto, Canada}, \bibinfo{pages}{1714--1729}.
\newblock
\urldef\tempurl%
\url{https://doi.org/10.18653/v1/2023.acl-long.96}
\showDOI{\tempurl}


\bibitem[Zhou et~al\mbox{.}(2023b)]%
        {zhou-etal-2023-case}
\bibfield{author}{\bibinfo{person}{Jinfeng Zhou}, \bibinfo{person}{Chujie Zheng}, \bibinfo{person}{Bo Wang}, \bibinfo{person}{Zheng Zhang}, {and} \bibinfo{person}{Minlie Huang}.} \bibinfo{year}{2023}\natexlab{b}.
\newblock \showarticletitle{{CASE}: Aligning Coarse-to-Fine Cognition and Affection for Empathetic Response Generation}. In \bibinfo{booktitle}{\emph{Proceedings of the 61st Annual Meeting of the Association for Computational Linguistics (Volume 1: Long Papers)}}, \bibfield{editor}{\bibinfo{person}{Anna Rogers}, \bibinfo{person}{Jordan Boyd-Graber}, {and} \bibinfo{person}{Naoaki Okazaki}} (Eds.). \bibinfo{publisher}{Association for Computational Linguistics}, \bibinfo{address}{Toronto, Canada}, \bibinfo{pages}{8223--8237}.
\newblock
\urldef\tempurl%
\url{https://doi.org/10.18653/v1/2023.acl-long.457}
\showDOI{\tempurl}


\end{thebibliography}



\end{CJK*}

\end{document}